\documentclass[11pt,a4paper]{article}
\usepackage{times,latexsym}
\usepackage{url}
\usepackage[T1]{fontenc}

\usepackage{tabularx}
\usepackage{amssymb,amsmath,amsthm,enumitem}
\usepackage{pifont}
\usepackage{tikz-dependency}
\usetikzlibrary{shapes.misc,automata,positioning}
\usepackage{verbatim}
\usepackage{float}
\usepackage{bm}
\usepackage{placeins}
\usepackage{booktabs} % nice tables
\usepackage{pgfplots}[inner frame sep=0]
\pgfplotsset{compat=1.17}
\DeclareMathSymbol{\shortminus}{\mathbin}{AMSa}{"39}
\usepackage{mathtools, nccmath}

\usepackage{array}
\usepackage{makecell}

\newcommand{\nlp}[0]{{\textsc{nlp}}}

\newcommand{\mlpq}[0]{\ensuremath{C_Q}}

\newcommand{\tm}[0]{{\textsc{rm}}}

\newcommand{\msize}[1]{\normalsize{#1}}
\newcommand{\level}[1]{{\msize{\textsc{#1}}}}

\newcommand{\pos}[0]{{\level{pos}}}

\newcommand{\synt}[0]{{\level{syn}}}
\newcommand{\back}[0]{{\level{back}}}

\newcommand{\actionPos}[1]{\textsc{pos}(#1)}

\newcommand{\actionReduce}[0]{\ensuremath
{\textsc{reduce}}}
\newcommand{\actionShift}[0]{\ensuremath{\textsc{shift}}}
\newcommand{\actionLeft}[1]{{$\textsc{left}_{#1}$}}
\newcommand{\actionRight}[1]{{$\textsc{right}_{#1}$}}

\newcommand{\actionBack}[0]{{\textsc{back}}}
\newcommand{\actionNoBack}[0]{{$\neg \textsc{back}$}}

\newcommand{\backArray}[0]{{$\textsc{ba}$}}
\newcommand{\wordIndex}[0]{{$wi$}}

\newcommand{\sul}[0]{{\textsc{sl}}}
\newcommand{\rl}[0]{{\textsc{rl}}}
\newcommand{\mdp}[0]{{\textsc{mdp}}}
\newcommand{\rlbt}[0]{{\textsc{rlb}}}

\usepackage[acceptedWithA]{tacl2021v1}

\title{Dependency Parsing with Backtracking using Deep Reinforcement Learning}

\author{Franck Dary, Maxime Petit, Alexis Nasr \\
  Aix Marseille Univ, Université de Toulon, CNRS, LIS, Marseille, France \\ \texttt{\{franck.dary,maxime.petit,alexis.nasr\}@lis-lab.fr}
}

\date{}

\begin{document}
\maketitle

\begin{abstract}
Greedy algorithms for NLP such as transition based parsing are prone to error propagation. One way to overcome this problem is to allow the algorithm to backtrack and explore an alternative solution in cases where new evidence contradicts the solution explored so far. In order to implement such a behavior, we use reinforcement learning and let the algorithm backtrack in cases where such an action gets a better reward than continuing to explore the current solution. We test this idea on both POS tagging and dependency parsing and show that backtracking is an effective means to fight against error propagation.
\end{abstract}

\section{Introduction}
\label{sec:introduction}

Transition based parsing has become a major approach in dependency parsing, since the work of \citet{yamada2003statistical} and \citet{nivre2004memory} for it combines linear time complexity and high linguistic performances.
The algorithm follows a local and greedy approach to parsing that consists in selecting at every step of the parsing process the action that maximizes a local score, typically computed by a classifier. The action selected is greedily applied to the current configuration of the parser and yields a new configuration.

At training time, an oracle function transforms the correct syntactic tree of a sentence into a sequence of correct (configuration, action) pairs. These pairs are used to train the classifier of the parser. The configurations that do not pertain to the set of correct configurations are never seen during training. 

At inference time, if the parser predicts and executes an incorrect action, it produces an incorrect configuration, with respect to the sentence being parsed, which might have never been seen during training, yielding a poor prediction of the next action to perform. Besides, the parser follows a single hypothesis by greedily selecting the best scoring action. The solution built by the parser can be sub-optimal for there is no guarantee that the sum of the scores of the actions selected maximizes the global score.

These are well known problems of transition based parsing and several solutions have been proposed in the literature to overcome them, they will be briefly discussed in Section~\ref{sec:relatedWork}. The solution we propose in this article consists in allowing the parser to backtrack. At every step of the parsing process, the parser has the opportunity to undo its $n$ previous actions to explore alternative solutions. The decision to backtrack or not is taken each time a new word is considered, before trying to process it, by giving the current configuration to a binary classifier, that will assign a score to the backtracking action. Traditional supervised learning is not suited to learn such a score, since the training data contains no occurrences of backtrack actions. In order to learn in which situation a backtrack action is worthy, we use reinforcement learning. During training, the parser has the opportunity to try backtracking actions and the training algorithm responds to this choice by granting it a reward. If the backtracking action is the adequate move to make in the current configuration, it will receive a positive reward and the parser will learn in which situation backtrack is adequate.

The work presented here is part of a more ambitious project which aims at modeling the eye movements during human reading. More precisely, we are interested to predict regressive saccades: eye movements that bring the gaze back to a previous location in the sentence. 

There is much debate in the psycholinguistic literature concerning the reasons of such eye movements \cite{lopopolo2019dependency}. Our position with respect to this debate is the one advocated by \citet{rayner1994regressive} for whom part of these saccades are linguistically motivated and happen in situations where the reader incremental comprehension of the sentence is misguided by an ambiguous sentence start, until reaching a novel word which integration will prove incompatible with the current understanding and will trigger a regressive saccade, as in garden path sentences. Our long time project is to model regressive saccades with backtracking actions. Although this work enters in this long term project, the focus of this article is on the {\sc nlp} aspects of this program and propose a way to implement backtracking in the framework of transition based parsing. We will just mention some preliminary studies on garden path sentences in the conclusion.

In order to move in the direction of a more cognitively plausible model, we add two constraints to our model. 

The first one concerns the text window around the current word that the parser takes into account when predicting an action. 
This window can be seen as an approximation of the sliding window introduced by \citet{mcconkie1975span} to model the perceptual span of a human reader\footnote{It is actually a rough approximation of the sliding window since it defines a span over words and not characters.}.
Transition based parsers usually allow taking into account the right context of the current word.
The words in the right context constitute a rich source of information and yield better predictions of the next action to perform.

In our model, the parser does not have access to this right context, simulating the fact that a human reader has only a limited access to the right context (few characters to the right of the fixation point~\cite{mcconkie1976asymmetry}). It is only after backtracking that a right context is available for it has been uncovered before backtrack took place.

The second constraint is incrementality. When performing several tasks, such as \pos{} tagging and parsing, as will be done in the tagparser described in Section~\ref{sec:model}, these tasks are organized in an incremental fashion. At each step, a word is read, \pos{} tagged and integrated in the syntactic structure, a more cognitively plausible behavior than a sequential approach where the whole sentence is first \pos{} tagged then parsed.

The structure of the paper is the following: in Section~\ref{sec:relatedWork}, we compare our work to other approaches in transition based parsing which aim at proposing solutions to the two problems mentioned above. In Section~\ref{sec:model}, we describe the model that we use to predict the linguistic annotation and introduce the notion of a \back{} action. In Section~\ref{sec:training} we show how backtracking actions can be predicted using Reinforcement Learning. Section~\ref{sec:experiments} describes the experimental part of the work and discusses the results obtained. Section~\ref{sec:conclusions} concludes the paper and presents different directions in which this work will be extended.

\section{Related Work}
\label{sec:relatedWork}

Several ways to overcome the two limits of transition based parsing mentioned in the introduction, namely training the parser with only correct examples and exploring only a single hypothesis at inference time have been explored in the literature.

\paragraph{Beam Search}
The standard solution to the single hypothesis search is beam search which allows considering a fixed number of solutions in parallel during parsing.
Beam search is a general technique that has been applied to transition based parsing in many works, among which, \citet{zhang2008tale}, \citet{huang2010dynamic} and \citet{zhang2012analyzing}.
They show that exploring a fixed number of solution increases the linguistic performances over a single hypothesis parser.
In this work, we do not use a beam search algorithm, but we do explore several solutions in a non parallel fashion, using backtracking.

\paragraph{Dynamic Oracles}
A first way to overcome the lack of exploration during training problem,
is the proposition by \citet{goldberg2012dynamic} to replace the standard oracle of transition based parsing by a {\it dynamic oracle} that is able to determine the optimal action $a$ to perform for an incorrect configuration $c$.
During training, the dynamic oracle explores a larger part of the configuration space than the static oracle and produces for an incorrect configuration $c$ an optimal action $a$.
The pair $(c,a)$ is given as training example to the classifier, yielding a more robust classifier that is able to predict the optimal action in some incorrect configurations.
\citet{ballesteros2016training} show that the principle of the dynamic oracle can be adapted to train the greedy Stack-LSTM dependency parser of \citet{dyer-etal-2015-transition}, improving its performances.

\citet{yu-etal-2018-approximate} describe a method, for training a small and efficient neural network model, that approximates a dynamic oracle for any transition system.
Their model is trained using reinforcement learning.

In this paper, we use dynamic oracles in two different ways.
First, as baselines for models allowing some exploration during training. Second, in the definition of the immediate reward function, as explained in Section~\ref{sec:training}.

\paragraph{Reinforcement Learning}
A second answer to the lack of exploration problem is reinforcement learning.
The exploitation, exploration trade-off of reinforcement learning allows, at training time, the model to explore some incorrect configurations and to learn a policy that selects in such cases the action that maximizes the long-term reward of choosing this action.
Reinforcement learning is well suited for transition based algorithms.

To the best of our knowledge, two papers directly address the idea of training a syntactic transition based parser with reinforcement learning: \citet{zhang2009dependency} and \citet{le2017tackling}.

\citet{zhang2009dependency} cast the problem, as we do, as a Decision Markov Process but use a Restricted Boltzmann Machine in order to compute scores of actions and use the SARSA algorithm to compute an optimal policy. In our case, we use deep Q-learning based on a Multi Layer Perceptron, as described in Section~\ref{sec:training}. In addition, their immediate reward function is based on the number of arcs in the gold tree of a sentence that are absent in the tree being built. We also use an immediate reward function, but it is based on the number of arcs in the gold tree that cannot anymore be built given the current tree being built, an idea introduced in the dynamic oracle of~\citet{goldberg2012dynamic}.

Two major differences distinguish our approach and the work of \citet{le2017tackling}. The first one is the idea of pre-training a parser in a supervised way and then fine-tuning its parameters using reinforcement learning. In our case, the parser is trained from scratch using reinforcement learning.
The reason for this difference is related to our desire to learn how to backtrack: it is difficult to make the parser learn to backtrack when it has been initially trained not to do it (using standard supervised learning). The second major difference is the use of a global reward function, that is computed after the parser has parsed a sentence. In our case, as mentioned above, we use an immediate reward. The reason for this difference is linked to pretraining. Since we do not pre-train our parser, and allow it to backtrack, granting a reward at the end of the sentence does not allow the parser to converge since reaching the end of the sentence is almost impossible using a non pre-trained parser. Other less fundamental differences distinguish our approach, such as the use of Q-learning, in our case, to find an optimal policy, where they use a novel algorithm called Approximate Learning Gradient, based on the fact that their model's output is a probability distribution over parsing actions (as in standard supervised learning). Another minor difference is the exploration of the search space in the training phase. They sample the next action to perform using the probability distribution computed by their model, while we use an adaptation of $\varepsilon$-greedy we describe in Section~\ref{sec:training}.

Reinforcement learning has also been used to train a transition based model for other tasks. \citet{naseem-etal-2019-rewarding} present a method to fine-tune the Stack-LSTM transition-based AMR parser of \citet{ballesteros-al-onaizan-2017-amr} with reinforcement learning, using the Smatch score of the predicted graph as reward.
For semantic dependency parsing, \citet{kurita-sogaard-2019-multi} found that fine-tuning their parser with policy gradient allows it to develop an easy-first strategy, reducing error propagation.

The fundamental difference between all papers cited above and our work is the idea of adding backtracking in a greedy transition based model. We use reinforcement learning as a mean to achieve the exploration necessary to this goal.
In term of parsing performance, our method will fare lower than the state of the art in transition based parsing for our parser is constrained to not see words beyond the current word, a constraint that comes from the fact that our long term goal is not to improve parsing performances but to find a natural way to encourage a parser to simulate regressive saccades observed during human reading.

\section{Backtracking Reading Machines}
\label{sec:model}

Our model is an extension of the Reading Machine, a general model for \nlp{} proposed in \citet{dary-nasr-2021-reading} that generalizes transition based parsing to other \nlp{} tasks. A Reading Machine is a finite automaton which states correspond to linguistic levels. There can be, for example, one state for \pos{} tagging, one state for lemmatization, one state for syntactic parsing\ldots When the machine is in a given state, an {\it action} is predicted, which generally writes on an output tape a label corresponding to the prediction just made\footnote{Some actions can be complex and change, for example, the state of an internal stack. All details concerning the Reading Machine can be found in \citet{dary-nasr-2021-reading}.}. There are usually as many output tapes as there are levels of linguistic predictions and at least one input tape which usually contains the words of the sentence to process\footnote{For the sake of simplicity, we consider in this paper that the text to process has already been segmented into sentences and tokenized into words, even if the Reading Machine allows performing these two operations.}. Predictions are realized by classifiers, that take as input the configuration of the machine and compute a probability distribution over the possible actions. Configurations are complex objects that describe the aspects of the machine that are useful in order to predict the next action to perform. Among the important elements of a configuration for the rest of the paper, we can cite the {\it current state} of the machine, the {\it word index}, noted \wordIndex{}, that is the position of the word currently processed and the {\it history}: a list of all actions performed so far.

The text is read word by word, a window of an arbitrary size centered on \wordIndex{} defines the part of the text that can be used to make predictions on the current word. 

\tikzset{every loop/.style={min distance=5mm,looseness=7}}

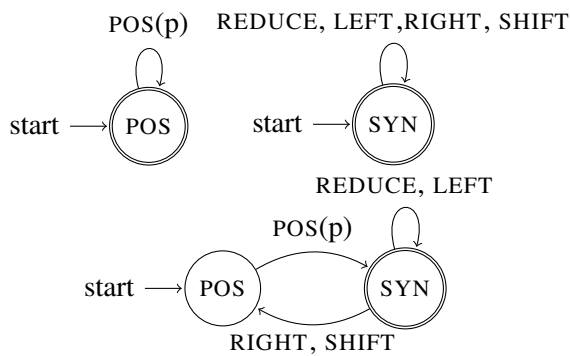
\begin{figure}[htb]
    \centering

\begin{tikzpicture}[shorten >=1pt,on grid,auto] 
   \node[state, initial, accepting] (P) [] {\pos}; 
    \path[->] 
%    (T) edge  node {} (P)
%        edge  [in=30,out=60,loop] node {} (T)
    (P) edge [loop above] node{\actionPos{p}} (P);
\end{tikzpicture} \begin{tikzpicture}[shorten >=1pt,on grid,auto] 
   \node[state, initial, accepting] (S) [right=2.4cm of P] {\synt};
    \path[->] 
%    (T) edge  node {} (P)
%        edge  [in=30,out=60,loop] node {} (T)
    (S) edge [loop above] node{\actionReduce{}, \actionLeft{},\actionRight{}, \actionShift{}} (S);
\end{tikzpicture}

\begin{tikzpicture}[shorten >=1pt,on grid,auto] 
   \node[state, initial] (P) [] {\pos}; 
   \node[state, accepting] (S) [right=2.4cm of P] {\synt};
    \path[->] 
%    (T) edge  node {} (P)
%        edge  [in=30,out=60,loop] node {} (T)
    (P) edge [bend left, above] node{\actionPos{p}} (S)
    (S) edge [bend left, below] node{\actionRight{}, \actionShift{}} (P)
    (S) edge [loop above] node{\actionReduce{}, \actionLeft{}} (S);
\end{tikzpicture}

    \caption{Three  simple reading machines. The top left machine performs \pos{} tagging, the right top one, unlabeled dependency parsing and the bottom one performs the two tasks simultaneously.}
    \label{fig:basic-architecture}
\end{figure}

Figure~\ref{fig:basic-architecture} shows the architecture of three simple machines. The two machines in the top part of the figure realize a single task. The machine on the left part realizes \pos{} tagging. It is made of a single state and has a series of transitions that loop on the state. The machine has one input tape from which words are read and one output tape, on which predicted \pos{} are written. Each transition is labeled with a tagging action of the form \actionPos{p} that simply writes the \pos{} tag p on the \pos{} output tape at the word index position.

The machine on the right implements an arc-eager transition based parser that produces unlabeled dependency trees. It has the same simple structure than the tagging machine.  Its transitions are labeled with the four standard arc-eager actions of unlabeled transition based parsing: \actionLeft{}, \actionReduce{}, \actionShift{} and \actionRight{}. The machine has two input tapes, one for words and one for \pos{} tags, and one output tape on which it writes the index of the governor of the current word when \actionLeft{} or \actionRight{} actions are predicted.

The machine on the bottom part of the figure, which we call a tagparser, realizes the two tasks simultaneously in an incremental fashion. When in state \pos{}, the machine tags the current word then control jumps to the parser in order to attach the current word to the syntactic structure built so far or to store it in a stack. Once this is done, control returns to state \pos{} to tag the next word. The reason why the transitions labeled \actionRight{} and \actionShift{} lead to state \pos{} is that these two actions increase the word index \wordIndex{} and initiate the processing of a new word.  The machine has one input tape for words and two output tapes, one for \pos{} tags and one for the governor position.

We augment the machines described above in order to allow them to undo some preceding actions. This ability relies on three elements: (a) the definition of a new action, called \actionBack{}, that undoes a certain number of actions (b) the history of the actions performed so far in order to decide which actions to undo and (c) the definition of undoing an action.

Undoing an action amounts to recover the configuration that existed before the action was performed. This is quite straightforward for tagging and parsing actions\footnote{In order to be undone, actions \actionReduce{} and \actionLeft{} also need to store stack elements that have been popped.}.

\tikzset{every loop/.style={min distance=5mm,looseness=7}}

% version originale, plus lisible
%\begin{comment}
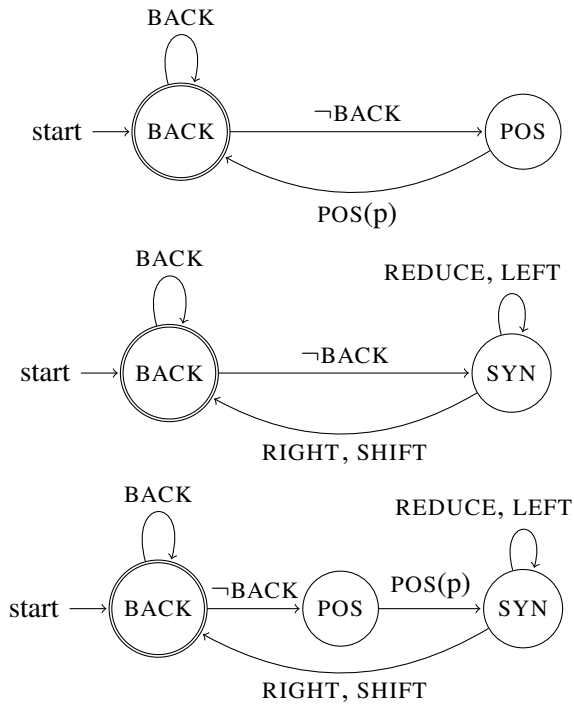
\begin{figure}[htb]
    \centering
\begin{tikzpicture}[shorten >=1pt,on grid,auto] 
   \node[state, initial, accepting] (B)  [] {\back}; 
   \node[state] (P) [right=4.5cm of B] {\pos}; 
    \path[->] 
    (B) edge  node{\actionNoBack{}}(P)
    (B) edge [loop above] node{\actionBack{}} (B)
    (P) edge [bend left, below] node{\actionPos{p}} (B);
\end{tikzpicture}

\begin{tikzpicture}[shorten >=1pt,on grid,auto] 
   \node[state, initial, accepting] (B)  [] {\back}; 
   \node[state] (S) [right=4.5cm of B] {\synt};
    \path[->] 
%    (T) edge  node {} (P)
%        edge  [in=30,out=60,loop] node {} (T)
    (B) edge  node{\actionNoBack{}}(S)
    (B) edge [loop above] node{\actionBack{}} (B)
    (S) edge [bend left, below] node{\actionRight{}, \actionShift{}} (B)
    (S) edge [loop above] node{\hspace*{-1.0cm}\actionReduce{}, \actionLeft{}} (B);
\end{tikzpicture}

\begin{tikzpicture}[shorten >=1pt,on grid,auto] 
   \node[state, initial, accepting] (B)  [] {\back}; 
   \node[state] (P) [right=2.4cm of B] {\pos}; 
   \node[state] (S) [right=2.4cm of P] {\synt};
    \path[->] 
%    (T) edge  node {} (P)
%        edge  [in=30,out=60,loop] node {} (T)
    (B) edge  node{\actionNoBack{}}(P)
    (B) edge [loop above] node{\actionBack{}} (B)
    (P) edge node{\actionPos{p}} (S)
    (S) edge [bend left, below] node{\actionRight{}, \actionShift{}} (B)
    (S) edge [loop above] node{\hspace*{-1.0cm}\actionReduce{}, \actionLeft{}} (B);
\end{tikzpicture}

    \caption{Three backtracking machines based on the machines of Figure\ref{fig:basic-architecture}}
    \label{fig:back-architecture}
\end{figure}

Three backtracking machines, based on the machines of Figure~\ref{fig:basic-architecture}, are shown in Figure~\ref{fig:back-architecture}. They all have an extra state, named \back{}, and two extra transitions. When in state \back{}, the machine predicts one of the two actions \actionBack{} or \actionNoBack{}. When action \actionNoBack{} is selected, control jumps either to state \pos{} or \synt{}, depending on the machine, and the machine behaves like the simple machines of Figure~\ref{fig:basic-architecture}. Action \actionNoBack{} does not modify the configuration of the machine. This is not the only possible architecture for backtracking machines, this point will be briefly addressed in the conclusion.

If action \actionBack{} is selected, the last actions of the history are undone until a \actionNoBack{} action is reached. This definition of action \actionBack{} allows undoing all actions that are related to the previous word. After \actionBack{} has been applied, the configuration of the machine is almost the same as the configuration it was in before processing the current word. There is however a major difference: it has now access to the following word. Otherwise, the machine would deterministically predict the  action it has predicted before. One can notice that the transition labeled \actionBack{} in the machine loops on a single state, this feature allows the machine to perform several successive \actionBack{} actions.

\begin{figure}[htb]
\center
{\footnotesize
%\begin{tabular}{|p{0.8cm}|p{0.8cm}|p{0.8cm}|p{0.8cm}|p{0.8cm}|} \hline
%&&&&\\ \hline
%{\sc } & {\sc} & {\sc} & {\sc} & {\sc} \\ \hline
%{\bf the} & {\sc} & {\sc} & {\sc} & {\sc} \\ \hline
%0 & 0 & 0 & 0 & 0\\ \hline
%\end{tabular}

%\actionNoBack{}, \actionPos{{\sc det}}, \actionShift{}

%\begin{tabular}{|p{0.8cm}|p{0.8cm}|p{0.8cm}|p{0.8cm}|p{0.8cm}|} \hline
%&&&&\\ \hline
%{\sc det} & {\sc } & {\sc} & {\sc} & {\sc} \\ \hline
%the & {\bf old} & {\sc} & {\sc} & {\sc} \\ \hline
%0 & 0 & 0 & 0 & 0\\ \hline
%\end{tabular}

%\actionNoBack{}, \actionPos{{\sc adj}}, \actionShift{}

\begin{tabular}{|p{0.8cm}|p{0.8cm}|p{0.8cm}|p{0.8cm}|p{0.8cm}|} \hline
 &  &&&\\ \hline
{\sc det} & {\sc adj} & {\sc } & {\sc} & {\sc} \\ \hline
the & old & {\bf man} & {\sc} & {\sc} \\ \hline
0 & 0 & 0 & 0 & 0\\ \hline
\end{tabular}

\actionNoBack{}, \actionPos{{\sc noun}}, \actionLeft{}, \actionLeft{}, \actionShift{}

\begin{tabular}{|p{0.8cm}|p{0.8cm}|p{0.8cm}|p{0.8cm}|p{0.8cm}|} \hline
3 & 3 &&&\\ \hline
{\sc det} & {\sc adj} & {\sc noun} & {\sc} & {\sc} \\ \hline
the & old & man & {\bf the} & {\sc} \\ \hline
0 & 0 & 0 & 0 & 0\\ \hline
\end{tabular}

\actionBack{}

\begin{tabular}{|p{0.8cm}|p{0.8cm}|p{0.8cm}|p{0.8cm}|p{0.8cm}|} \hline
$-$ & $-$ &&&\\ \hline
{\sc det} & {\sc adj} & $-$ & {\sc} & {\sc} \\ \hline
the & old & {\bf man} & the & {\sc} \\ \hline
0 & 0 & 0 & 0 & 0\\ \hline
\end{tabular}

\actionBack{}

\begin{tabular}{|p{0.8cm}|p{0.8cm}|p{0.8cm}|p{0.8cm}|p{0.8cm}|} \hline
 $-$& $-$ &&&\\ \hline
{\sc det} & $-$ & $-$ & {\sc} & {\sc} \\ \hline
the & {\bf old} & man & the & {\sc} \\ \hline
0 & 0 & 1 & 1 & 0\\ \hline
\end{tabular}

\actionNoBack{}, \actionPos{{\sc noun}}, \actionLeft{}, \actionShift{}

\begin{tabular}{|p{0.8cm}|p{0.8cm}|p{0.8cm}|p{0.8cm}|p{0.8cm}|} \hline
 2& $-$ &&&\\ \hline
{\sc det} & {\sc noun} & $-$ & {\sc} & {\sc} \\ \hline
the & old & {\bf man} & the & {\sc} \\ \hline
0 & 0 & 1 & 1 & 0\\ \hline
\end{tabular}

\actionNoBack{}, \actionPos{{\sc verb}}, \actionLeft{}, \actionShift{}

\begin{tabular}{|p{0.8cm}|p{0.8cm}|p{0.8cm}|p{0.8cm}|p{0.8cm}|} \hline
 2& 3 &&&\\ \hline
{\sc det} & {\sc noun} & {\sc verb} & {\sc} & {\sc} \\ \hline
the & old & man & {\bf the} & {\sc} \\ \hline
0 & 0 & 1 & 1 & 0\\ \hline
\end{tabular}

\actionNoBack{}, \actionPos{{\sc det}}, \actionShift{}

%\begin{tabular}{|p{0.8cm}|p{0.8cm}|p{0.8cm}|p{0.8cm}|p{0.8cm}|} \hline
% 2& 3 &&&\\ \hline
%{\sc det} & {\sc noun} & {\sc verb} & {\sc det} & {\sc} \\ \hline
%the & old & man & the & {\bf boat} \\ \hline
%0 & 0 & 1 & 1 & 0\\ \hline
%\end{tabular}

%\actionNoBack{}, \actionPos{{\sc noun}}, \actionLeft{}, \actionRight{}

%\begin{tabular}{|p{0.8cm}|p{0.8cm}|p{0.8cm}|p{0.8cm}|p{0.8cm}|} \hline
% 2& 3 &&5&3\\ \hline
%{\sc det} & {\sc noun} & {\sc verb} & {\sc det} & {\sc noun} \\ \hline
%the & old & man & the & {\bf boat} \\ \hline
%0 & 0 & 1 & 1 & 0\\ \hline
%\end{tabular}

}

    \caption{Processing the garden path sentence {\it the old man the boat} with a bactracking tagparser. After reading the second determiner, the machine backtracks in order to reanalyse words {\it old} and {\it man}. }
    \label{fig:theOldManTheBoat}
\end{figure}

Figure~\ref{fig:theOldManTheBoat} shows how the tagparsing machine of Figure~\ref{fig:back-architecture} would ideally process the sentence {\it the old man the boat}, a classical garden path sentence for which, two words ({\it old} and {\it man}) should be re-analysed after the noun phrase {\it the boat} has been read. The figure describes the machine configuration each time it is in state \back{}. Three tapes are represented: the input tape, that contains tokens, the \pos{} tape and the parsing tape that contains the index of the governor of the current word. The figure also represents, at the bottom, the \backArray{} array, that is described below, as well as the sequence of actions predicted since the last visit to state \back{}. The current word appears in boldface. The figure shows two successive occurrences of a \back{} actions, after the second determiner is read, leading to the re-analysis of the word {\it old} that was tagged {\sc adj} and the word {\it man} that was tagged {\sc noun}.

A backtracking machine as the one described above can run into infinite loops: nothing prevents it to repeat endlessly the same sequence of actions. One can hope that, during training, such a behavior leads to poor performances and is ruled out. But there is no guarantee that this will be the case. Furthermore, we would like to prevent the machine for exploring, at inference time, the whole (or a large part of the) configuration space. In order to do so, we introduce a constraint on the number of times a \actionBack{} action is taken when parsing a sentence. A simple way to introduce such a constraint is to limit to a given constant $k$ the number of authorized \actionBack{} actions per word. This feature is implemented by introducing an array \backArray{} of size $n$, where $n$ is the number of words of the sentence to process. Array \backArray{} is initialized with zeros, and every time a \actionBack{} action is predicted in position $i$, the value of \backArray{}[\wordIndex{}] is incremented. When the machine is in state \back{} and \backArray{}[\wordIndex{}] is equal to $k$, performing a \actionBack{} action is not permitted. A \actionNoBack{} action is forced, bypassing the classifier that decides whether to backtrack or not. 

The introduction of array \backArray{} and the parameter $k$ defines an upper bound on the size of the action sequence for a sentence of length $n$. This upper bound is equal to $3nk+2n$ for the tagger, $4nk+3n$ for the parser and $5nk+4n$ for the tagparser\footnote{See Appendix A for details.}. As one can notice, linearity is preserved. In our experiments, we chose $k=1$.
\section{Training}
\label{sec:training}

Reading Machines, as introduced by \citet{dary-nasr-2021-reading} are trained in a supervised learning fashion. Given data annotated at several linguistic levels, an oracle function decomposes it into a sequence of configurations and actions $(c_0, a_0, c_1, a_1, \dots, c_n, a_n)$. This sequence of configurations and actions constitute the training data of the classifiers of the machine to train: pairs $(c_i, a_i)$ are presented iteratively to the classifier during the training stage.  A backtracking Reading Machine cannot be trained this way since there are no occurrences of \actionBack{} actions in the data. In order to learn useful occurrences of such actions, the training process should have the ability to generate some \actionBack{} actions and be able to measure if this action was beneficial.

In order to implement such a behavior, we use Reinforcement Learning (\rl{}). We cast our problem as a Markov Decision Process (\mdp{}). In an \mdp{}, an agent (the parser) is in configuration $c_t$ at time $t$. It selects an action $a_t$ from an action set (made of the tagging actions, the parsing actions and the \actionBack{} and \actionNoBack{} actions) and performs it. We note $C$ the set of all configurations and $A$ the set of actions. The environment (the annotated data) responds to action $a_t$ by giving a reward $r_t = r(c_t, a_t)$ and by producing the succeeding configuration $c_{t+1} = \delta(c_t, a_t)$. In our case, configuration $c_{t+1}$ is deterministically determined by the structure of the \tm{}. The reward function gives high reward to actions that move the parser towards the correct parse of the sentence. The fundamental difference with supervised training is that, during training, the agent is not explicitly told which actions to take, but instead must discover which action yields the most reward through trial and error. This feature gives the opportunity for the parser to try some \actionBack{} actions, provided that a reward can be computed for such actions.

Given an \mdp{} that indicates the reward associated to applying action $a$ in configuration $c$, the goal is to learn a function $q_*(c,a)$ which maximizes the total amount of reward (also called discounted return) that can be expected after action $a$ has been applied. Once this function (or an approximation of it) is computed, one can use it to select which action to choose when in configuration $c$, by simply picking the action that maximizes function $q_*$. Such a behavior is called an optimal policy in the \rl{} literature. 
Function $q_*$ is the solution to the Bellman optimality equation\footnote{This is actually a simplified form of the Bellman optimality equation, due to the fact that our \mdp{} is deterministic: applying action $a$ in configuration $c$ yields configuration $\delta(c,a)$ with probability $1$.}:\useshortskip
\begin{equation*}
q_*(c,a) = r(c,a) + \gamma \max_{a'}q_*(\delta(c,a), a')
\end{equation*}
where $\gamma \in [0,1]$ is the discount factor that allows discounting the reward of future actions. The equation expresses the relationship between the value of an action $a$ in configuration $c$ and the value of the best action $a'$ that can be performed in its successor configuration $\delta(c,a)$. This recursive definition is the basis for algorithms that iteratively approximate $q_*$, among which Q learning \cite{watkins1989learning}, which approximates $q_*$ with a function called $Q$. In Q learning, during training, each time an action $a$ is chosen in configuration $c$, the value of $Q(c,a)$ is updated:\useshortskip
\begin{equation*}
Q(c,a) \leftarrow Q(c,a) +  \alpha( Q'(c,a) - Q(c,a))
\end{equation*}
where $\alpha$ is a learning rate and $Q'(c,a)$ is a new estimation of $Q(c,a)$:\useshortskip
\begin{equation*}
Q'(c,a) = r(c,a) + \gamma \max_{a'} Q(\delta(c,a),a')
\end{equation*}

It has been proven~\cite{watkins1992q} that such iterative algorithm converges towards the $q_*$ function.

In order to store the values of $Q$, the algorithm uses a large table that has an entry for every (configuration, action) pair.
In our case, there are far too many configurations to allocate such a table. Instead we use a simple form of deep Q learning~\cite{mnih2013playing} and approximate function $Q$ using a multi-layered perceptron \mlpq{}.

\mlpq{} takes as input a configuration $c$ and outputs a vector whose dimension is the number of different actions in the action set. The component of the vector corresponding to action $a$ is the approximation of $Q(c,a)$ computed by \mlpq{}. It is noted $\mlpq{}(c,a)$.

During training, every time an action $a$ is performed by the parser in configuration $c$, the parameters of \mlpq{} are updated using gradient descent of the loss function. The loss function should be defined in a way to  minimize the difference between the actual value $\mlpq(c,a)$ and its updated value $\mlpq'(c,a)$. This difference is computed with the smooth $l1$ loss~\citep{girshick2015fast}:\useshortskip
\begin{equation*}
L(c,a) = l1(\mlpq{}(c,a), \mlpq'{}(c,a))
\end{equation*}

where $\mlpq'{}(c,a)$ is computed as follows:\useshortskip
\begin{equation*}
\mlpq'{}(c,a) = r(c,a) + \gamma \max_{a'} \mlpq{}(\delta(c,a), a')
\end{equation*}

\paragraph{Reward Functions}

In \rl{}, the training process is guided by the reward $r(c,a)$ granted by the environment when action $a$ is performed in configuration $c$. Defining a reward function for tagging action and parsing action is quite straightforward. 

In the case of tagging, the reward should be high when the tag chosen for the current word is correct and low when it is not. A simple reward function is one that associates, for example, value $0$ in the first case and $\shortminus1$ in the second. More elaborate reward functions could be defined that penalize more some confusions (for example tagging a verb as a preposition).

In the case of parsing, a straightforward reward function will give a reward of zero for a correct action and a negative reward for an incorrect one. We use a slightly more complex function inspired by the dynamic oracle of \citet{goldberg2012dynamic}. This function, in the case of an incorrect action $a$, counts the number of dependencies of the correct analysis that cannot anymore be predicted due to $a$. The reward for $a$ is the opposite of this number. Actions that cannot be executed, such as popping an empty stack, are granted a reward of $\shortminus1.5$.

Defining the reward function for \actionBack{} actions is more challenging. When a \actionBack{} action is performed, a certain number of actions $a_i \dots a_{i+k}$ are undone. Each of these actions was granted a reward $r_i \dots r_{i+k}$. Let's call $E = -\Sigma_{t=0}^k r_{i+t}$ the opposite of the sum of these rewards ($E\geq0$). The larger $E$ is, the more errors have been made. Let's call $\varphi(E)$ the function that computes the reward for executing a \actionBack{} action, given $E$.
Formally, we want $\varphi$ to respect the following principles:

\begin{enumerate}
    \item Don't execute a \actionBack{} action if there are no errors: $\varphi(0) < 0$.
    \item $\varphi(E)$ should be increasing with respect to $E$: the more errors, the more a \actionBack{} action should be encouraged.
    \item $\varphi(E)$ should not grow too fast with respect to $E$. Granting too much reward to \actionBack{} actions encourages the system to make errors in order to correct them with a highly rewarded \actionBack{} action.
\end{enumerate}

A function that fulfills these principles is the following:
\[   \varphi(E) = \left\{
\begin{array}{ll}
      -1 & \ \textit{if} \ E = 0 \\
      \ln(E+1) & \textit{else}\\
\end{array} 
\right. \]

It is the function that we use to compute the reward of a \actionBack{} action.

\paragraph{Exploring the Configurations Space}

\begin{figure}[htb]
\centering
\begin{tikzpicture}
    \begin{axis}[
        xlabel=Epoch,
        scale only axis=true,
        xmin=1,
        xmax=20,
        ymin=0,
        ymax=1,
        width=7.0cm,
        height=3.0cm,
        no markers,
        every axis plot/.append style={ultra thick},
        legend style={at={(1,0.75)},anchor=north east}
        ]
    \addplot[smooth,blue] plot coordinates {
(1,0.6)
(2,0.49)
(3,0.4)
(4,0.34)
(5,0.28)
(6,0.24)
(7,0.21)
(8,0.19)
(9,0.17)
(10,0.15)
(11,0.14)
(12,0.13)
(13,0.12)
(14,0.12)
(15,0.12)
(16,0.11)
(17,0.11)
(18,0.11)
(19,0.11)
(20,0.1)
    };
    \addlegendentry{~~$\varepsilon$ (random)}
    \addplot[smooth,color=orange]
        plot coordinates {
(1,0.3)
(2,0.18)
(3,0.11)
(4,0.07)
(5,0.04)
(6,0.02)
(7,0.01)
(8,0.01)
(9,0.01)
(10,0.0)
(11,0.0)
(12,0.0)
(13,0.0)
(14,0.0)
(15,0.0)
(16,0.0)
(17,0.0)
(18,0.0)
(19,0.0)
(20,0.0)
        };
    \addlegendentry{$\beta$ (oracle)}
    \addplot[smooth,color=green]
        plot coordinates {
(1,0.10000000000000009)
(2,0.33000000000000007)
(3,0.49)
(4,0.59)
(5,0.6799999999999999)
(6,0.74)
(7,0.78)
(8,0.8)
(9,0.82)
(10,0.85)
(11,0.86)
(12,0.87)
(13,0.88)
(14,0.88)
(15,0.88)
(16,0.89)
(17,0.89)
(18,0.89)
(19,0.89)
(20,0.9)
        };
    \addlegendentry{~~~Exploitation}
    \end{axis}
    \pgfresetboundingbox
    \path
    (current axis.south west) -- ++(-0.1in,-0.25in)
    rectangle (current axis.north east) -- ++(-0.1in,-0.1in);
    \end{tikzpicture}
    \caption{Probabilities of choosing the next action during RL. At random, following the oracle or the model.}
    \label{fig:probas}
\end{figure}
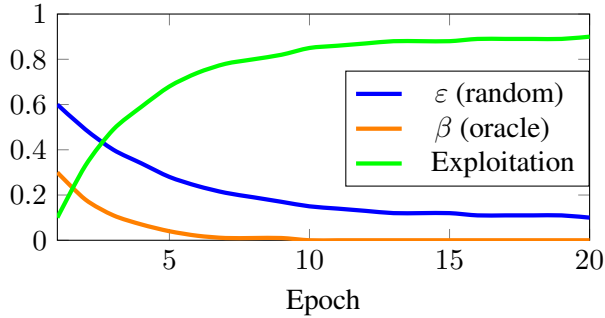

In order to learn useful \actionBack{} actions, the parser should try, in the training phase, to perform some \actionBack{} actions and update the classifier based on the reward it receives from the environment. One standard way to do that is to adopt an $\varepsilon$-greedy policy in which the model selects a random action with probability $\varepsilon$ or the most probable action as predicted by the classifier with probability $1- \varepsilon$. This setup allows the system to perform exploitation (choosing the best action so far) as well as exploration (randomly choosing an action). We adopt a variant of this policy, based on two parameters $\varepsilon$ and $\beta$ with $0 \leq \varepsilon \leq 1$, $0 \leq \beta \leq 1$ and $\varepsilon + \beta \leq 1$.
As in standard $\varepsilon$-greedy policy, the agent chooses a random action with probability $\varepsilon$, it chooses the correct action as given by the oracle\footnote{\actionNoBack{} in the case of the back state.} with probability $\beta$ and finally, it chooses the most probable action as predicted by the classifier with probability $1-(\varepsilon + \beta)$.
Parameter $\beta$ has been introduced in order to speed up training. In the beginning of the training process, the system is encouraged to follow the oracle. Then, as training progresses, the system relies more on its predicting capacity (exploitation augments) and less on the oracle.
Figure~\ref{fig:probas} shows the evolution of theses parameters.

\section{Experiments}
\label{sec:experiments}

Three machines were used in our experiments: a tagger, a parser and a tagparser, based on the architectures of Figure~\ref{fig:back-architecture}. Each of the three machines has been trained in three different learning regimes: Supervised Learning (\sul{}) using a dynamic oracle, Reinforcement Learning without back actions (\rl{}) and  Reinforcement Learning with back actions (\rlbt{}).

\subsection{Universal Dependencies Corpus}

Our primary experiments were conducted on a French Universal Dependencies corpus \citep{ud}, more specifically the GSD corpus, consisting of $16,341$ sentences and $400,399$ words.
The original split of the data was $88\%$ train, $9\%$ dev and $3\%$ test.
The size of the test set being too small to obtain significant results, we decided to use a $k$-fold strategy, where all the data was first merged, randomly shuffled and then split into ten folds, each fold becoming the test set of a new train/dev/test split of the data in the following proportions: $80\%/10\%/10\%$.

Using the ten folds was unnecessary to get significant results, we therefore decided to limit ourselves to three folds. The size of the test set for which results are reported has a size of $4,902$ sentences and $124,560$ words.

\subsection{Experimental Setup}

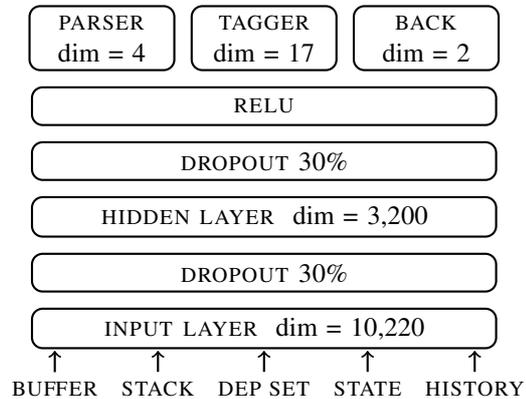
\begin{figure}[htb]
\small
\centering
\begin{tikzpicture}
\tikzset{layer/.style={text width=5.87cm,minimum height=0.5cm,align=center,draw, rounded corners,thick}}
\tikzset{decision/.style={text width=1.67cm,align=center,draw,rounded corners,thick}}
\node (1) [layer] {\textsc{input layer}~~dim~=~10,220};
\node (2) [above=0.2cm of 1, layer] {\textsc{dropout} 30\%};
\node (3) [above=0.2cm of 2, layer] {\textsc{hidden layer}~~dim~=~3,200};
\node (4) [above=0.2cm of 3, layer] {\textsc{dropout} 30\%};
\node (5) [above=0.2cm of 4, layer] {\textsc{relu}};
\node (6) [above=0.2cm of 5, decision] {\textsc{tagger}\\dim~=~17};
\node (7) [left=0.2cm  of 6, decision] {\textsc{parser}\\dim~=~4};
\node (8) [right=0.2cm of 6, decision] {\textsc{back}\\dim~=~2};
\node (9) [below=0.3cm of 1, align=center] {\textsc{dep set}};
\node (10) [left=0.05cm of 9, align=center] {\textsc{stack}};
\node (11) [left=0.05cm of 10, align=center] {\textsc{buffer}};
\node (12) [right=0.05cm of 9, align=center] {\textsc{state}};
\node (13) [right=0.05cm of 12, align=center] {\textsc{history}};
\path[->,thick]
(9) edge ++(0,0.5cm)
(10) edge ++(0,0.5cm)
(11) edge ++(0,0.5cm)
(12) edge ++(0,0.5cm)
(13) edge ++(0,0.5cm);
\end{tikzpicture}
\caption{Structure of the classifier.}
\label{fig:classifier}
\end{figure}

Each machine consists of a single classifier, a Multi Layer Perceptron, with a single hidden layer of size $3200$. When the machine realizes several tasks, as in case of the tagparser, the classifier has one decision layer for each task. The output size of each decision layer is the number of actions of its corresponding task. A dropout of $30\%$ is applied to the input vector and to the output of the hidden layer. The output of the hidden layer is given as input to a ReLU function. The structure of the classifier has been represented in Figure~\ref{fig:classifier}.

The details of the features extracted from configurations and their encoding in the input layer of the classifier are detailed in Appendix B.

In the case of Supervised Learning, the machines are trained with a dynamic oracle.
In the beginning of the training process, the machines are trained with a static oracle, for two epochs. Then every two epoch, the machines are used to decode the training corpus and for each configuration produced (which could be incorrect) the dynamic oracle selects the optimal action and these (configuration, action) pairs are used to train the classifier.

Training the machines in the \rl{} regime is longer than training them in the \sul{} regime. In the first case, $200$ epochs were needed and $300$ in the second.
This difference is probably due to the larger exploration of the configuration space.

\subsection{Results - Performances}

\begin{table}[htb]
\centering
\begin{tabular}{lrrrr} \toprule
\multicolumn{1}{c}{}& \multicolumn{2}{c}{TAGGER} & \multicolumn{2}{c}{PARSER}\\ 
Regime & UPOS & $p$~val. & UAS & $p$~val.\\ \midrule
\rlbt{} & \textbf{97.65} & 0.000 & \textbf{88.21} & 0.000\\
\rl{} & 96.84 & 0.000 & 86.60 & 0.037\\
\sul{} & 96.11 & \_ & 86.17 & \_\\ \bottomrule
\addlinespace[5.0mm] \toprule
\multicolumn{1}{c}{}& \multicolumn{4}{c}{TAGPARSER} \\
Regime & UPOS & $p$~val. & UAS & $p$~val.\\ \midrule
\rlbt{} & \textbf{97.06} & 0.000 & \textbf{87.85} & 0.001 \\
\rl{} & 96.73 & 0.090 & 87.12 & 0.211  \\
\sul{} & 96.59 & \_ & 86.94 & \_  \\ \bottomrule
\end{tabular}
\caption{Performances of tagger, parser and tagparser under three learning regimes on our French corpus.}
\label{tab:res_ud}
\end{table}

The results for the three machines, under the three learning regimes are shown in Table~\ref{tab:res_ud}.
\pos{} tagging performances are measured with accuracy and displayed in column \texttt{UPOS}. Dependency parsing performances are measured with the unlabeled accuracy score (ratio of words that have been attached to the correct governor) and displayed in the \texttt{UAS} column. The $p$-value next to each score is a confidence metric indicating if the score is significantly better than the one below (that's why the last line is never given a $p$-value). This $p$-value has been estimated with a paired bootstrap resampling algorithm~\citep{koehn-2004-statistical}, using the script \citep{popel2017udapi} of the CoNLL 2018 shared task.

The table shows the same pattern for the three machines: the \rlbt{} regime gets higher results than the simple \rl{} regime which is itself better than the \sul{} regime. Two important conclusions can be drawn from these results. The first one is that \rlbt{} regime is consistently better than \sul{}: backtracking machines make less errors than machines trained in supervised mode. At this point we do not know whether this superiority comes from reinforcement learning or the addition of a \back{} action. In fact, previous experiments in \citet{zhang2009dependency} and \citet{le2017tackling} showed that reinforcement learning (without backtracking) can lead to better results than supervised learning. The comparison of \rlbt{} and \rl{} shows that most of the performance boost comes from backtracking.

The results of Table~\ref{tab:res_ud} also show that the tagparser gets better results than single task machines (the tagger and the parser) when trained with supervised learning.
Note that this comparison is possible because the parser was not given gold PoS as input, but instead the ones predicted by the tagparser.
These results are in line with the work of \citet{bohnet2012transition} and \citet{alberti2015improved} that show that joint prediction of  \pos{} tags and syntactic tree improves the performances of both. However, this is not true when the machines are trained with reinforcement learning. In this case the parser and the tagger get better results than the tagparser. One reason that could explain this difference is the size of the configuration space of the tagparser that is an order of magnitude larger than those of the tagger or the parser. We will return to this point in the conclusion.

\subsection{Results - Statistics}

\begin{table}[htb]
\centering
\tabcolsep=1.5mm
\begin{tabular}{@{}lrrrrrrrrr@{}} \toprule
 & PARSER & TAGGER & TAGPARSER \\
\midrule
\#Actions & 115,588 & 79,620 & 153,764 \\
\#Errs & 3,597 & 1,323 & 6,249 \\
\#Backs & 1,063 & 891 & 4,491 \\
bPrec & 76.86\% & 68.46\% & 73.48\% \\
bRec & 24.52\% & 46.49\% & 61.72\% \\
C$\rightarrow$C & 18.07\% & 28.65\% & 56.89\% \\
E$\rightarrow$E & 41.39\% & 26.09\% & 23.46\% \\
C$\rightarrow$E & 05.15\% & 02.79\% & 02.81\% \\
E$\rightarrow$C & 35.39\% & 42.47\% & 16.83\% \\ \bottomrule
\end{tabular}
\caption{Behavior comparison of three \rlbt{} machines.}
\label{tab:stats_back}
\end{table}

One can gain a better understanding of the effect of the \back{} actions performed by the three machines with the statistics displayed in Table~\ref{tab:stats_back}. Each column of the table concerns one machine trained in \rlbt{} mode. The first line shows the total number of actions predicted while decoding the test set, the second one, the number of errors made and the third one, the number of \back{} actions predicted. Lines four and five give the precision and recall of the \back{} actions. The precision is the ratio of \back{} actions that were predicted after an error was made while the recall is the ratio of errors after which a \back{} action was predicted. These two figures measure the error detection capabilities of the \back{} actions prediction mechanism. In the case of the parser, the precision is equal to $76.86\%$, which means that $76.86\%$ of the \back{} actions were predicted after an error was made and $24.52\%$ (recall) of the errors provoked a \back{} action prediction. The recall constitutes an upper bound of the errors that could be corrected. The four last lines break down the \back{} actions predicted into four categories. \texttt{C$\rightarrow$C} is the case where a \back{} action was predicted after a correct action, but did not change the action, \texttt{E$\to$E} is the case where a \back{} action was predicted after an error, but the error was not corrected, either the same erroneous action was predicted
or another erroneous one was predicted. \texttt{E$\to$C} is the case where a \back{} action was predicted after an error and has corrected it, while \texttt{C$\to$E} is the case where a correct action was replaced by an incorrect one after a \back{} action. 

Several conclusions can be drawn from these statistics. 

First, backtracking corrects errors. For the three machines, there are much more cases where an error is corrected rather than introduced after a \back{} action was predicted (\texttt{E$\rightarrow$C} $\gg$ \texttt{C$\rightarrow$E}). This means that the difference in scores that we observed between \rl{} and \rlbt{} in Table~\ref{tab:res_ud} can indeed be attributed to backtracking.

Second, backtracking is conservative. The number of predicted \back{} actions is quite low (around $1\%$ of the actions for the tagger and the parser and around $3\%$ for the tagparser) and the precision is quite high. The machines do not backtrack very often and they usually do it when errors are made. This is the kind of behavior we were aiming for. It can be modified by changing the reward function $\varphi$ of the \back{} action.

Third, tagging errors are easier to correct than parsing errors. The comparison of columns two and three (parser and tagger) shows that the tagger has a higher recall than the parser, tagging errors are therefore easier to detect. This comparison also shows that \texttt{E$\rightarrow$C} is higher for the tagger than it is for the parser, tagging errors are therefore easier to correct.

At last, the poor performances of the tagparser do not come from the fact that it does not backtrack. It actually does backtrack around three times as much as the parser or the tagger. But it has a hard time correcting the errors, most of the time, it reproduces errors made before. We will return to this point in the conclusion. 

\subsection{Results on Other Languages}

\begin{table}[htb]
\centering
\setlength{\tabcolsep}{5.5pt}
\begin{tabular}{@{}llrrr@{}} \toprule
Lang. & Corpus & Train & Dev & Test\\ \midrule
\sc{ar}& PADT       & 254,400 & 34,261 & 32,132\\
\sc{zh}& GSD       & 98,616 &12,663 & 12,012\\
\sc{en}& GUM       & 81,861 & 15,598 & 15,926\\
\sc{fr}& GSD        & 364,349 & 36,775 & 10,298\\
\sc{de}& HDT        & 2,753,627 & 319,513 & 326,250 \\
\sc{ro}& RRT      & 185,113 & 17,074 & 16,324\\
\sc{ru}& SynTag & 871,526 & 118,692 & 117,523\\ \bottomrule
\end{tabular}
\caption{Corpora used for experiments on new languages, with the size of training, development and test sets (in tokens).}
\label{tab:corpora}
\end{table}

In order to study the behavior of backtracking on other languages, we have trained and evaluated our system on six other languages from various typological origin: Arabic ({\sc ar}), Chinese ({\sc zh}), English ({\sc en}), German ({\sc de}), Romanian ({\sc ro}) and Russian ({\sc ru}). The experimental setup for these languages is different from the one we have used for French: we have used the original split in train, development and test sets, as defined in the Universal Dependencies corpora. We report in Table~\ref{tab:corpora} the corpora used for each language as well as the size of the training, development and test sets. We did not run experiments on the tagparser for it gave poor results on our experiments on French. Besides, for sanity check, we have rerun experiments on French data, using the original split in order to make sure that the difference of experimental conditions did not yield important differences in the results. The results of these experiments can be found in Table~\ref{tab:resMultiling}. The table indicates $p$-values of the difference between one system and the next best performing one. The system with the worse performances is therefore not associated to a $p$-value.

The results obtained on French are lower than the results obtained using the $k$-fold strategy. But the drop is moderate: $0.13\%$ for the tagger and of $0.52\%$ for the parser and could be explained simply by the difference of the test corpora on which the systems were evaluated.

\begin{table}[htb]
\centering
\begin{tabular}{lrrrr} \toprule
& \multicolumn{2}{c}{TAGGER} & \multicolumn{2}{c}{PARSER} \\
Regime & UPOS & $p$~val. & UAS & $p$~val.  \\ \midrule
\multicolumn{5}{c}{English-GUM} \\
\rlbt{} & \textbf{94.99} & 0.00 & \textbf{79.96} & 0.00 \\
\rl{} & 93.53 & 0.00 & 72.97 &  \\
\sul{} & 92.63 &  & 73.12 & 0.43 \\ \midrule
\multicolumn{5}{c}{French-GSD} \\
\rlbt{} & \textbf{97.53} & 0.00 & \textbf{87.70} & 0.17 \\
\rl{} & 96.64 & 0.10 & 86.63 &  \\
\sul{} & 96.25 &  & 86.97 & 0.33 \\ \midrule
\multicolumn{5}{c}{German-HDT} \\
\rlbt{} & \textbf{97.88} & 0.00 & \textbf{93.00} & 0.00 \\
\rl{} & 97.29 & 0.47 & 91.26 &  \\
\sul{} & 97.28 &  & 91.31 & 0.35 \\ \midrule
\multicolumn{5}{c}{Romanian-RRT} \\
\rlbt{} & \textbf{97.07} & 0.00 & \textbf{85.40} & 0.26 \\
\rl{} & 96.28 & 0.03 & 84.70 &  \\
\sul{} & 95.77 &  & 85.00 & 0.32 \\ \midrule
\multicolumn{5}{c}{Russian-SynTagRus} \\
\rlbt{} & \textbf{98.41} & 0.00 & \textbf{86.59} & 0.00 \\
\rl{} & 97.93 & 0.01 & 85.25 &  \\
\sul{} & 97.80 &  & 85.27 & 0.46 \\ \midrule
\multicolumn{5}{c}{Chinese-GSD} \\
\rlbt{} & \textbf{93.01} & 0.00 & \textbf{71.70} & 0.00 \\
\rl{} & 91.61 & 0.23 & 64.63 &  \\
\sul{} & 91.30 &  & 65.79 & 0.13 \\ \midrule
\multicolumn{5}{c}{Arabic-PADT} \\
\rlbt{} & \textbf{96.43} & 0.16 & 83.81 & 0.47 \\
\rl{} & 96.20 & 0.03 & 83.77 &  \\
\sul{} & 95.74 &  & \textbf{83.95} & 0.38 \\ \bottomrule
\end{tabular}
\caption{Results for seven languages.}
\label{tab:resMultiling}
\end{table}

We observe more or less the same pattern for the new languages: the highest performances are reached by reinforcement learning with backtrack (\rlbt{}), for both the tagger and the parser. The second best performing systems for tagging are usually trained with reinforcement learning but differences are usually non significant. In the case of the parser, the second best performing systems are trained in a supervised regime, but as was the case for tagging, the differences are often non significant. The performances on Arabic are different, where no significant advantage was observed when using backtracking. The reason for this is the agglutinative nature of Arabic and the tokenization conventions of UD that tokenizes agglutinated pronouns. The effect of this tokenization is to increase the distances between content words. The most common pattern that triggers a backtrack in tagging consists in going back to the previous word in order to modify its part of speech. In the case of Arabic, if the target of the backtrack has an agglutinated pronoun, the tagger has to perform two successive \back{} actions to realize the correction, a pattern that is more difficult to learn.

The general conclusions that we can draw therefore is that reinforcement learning with backtrack yields the best performances for both the parser and the tagger (with the exception of Arabic), but there are no notable differences between supervised learning with a dynamic oracle and reinforcement learning (without backtrack). 

\begin{table}[htb]
\begin{tabular}{@{}lrrrr@{}} \toprule
Lang	         &	\textbf{\sc en}      &	\textbf{\sc de}      &	\textbf{\sc ru}      &	\textbf{\sc ar}      \\ \midrule
\#Act	         &	31,852  &	652,500 &	234,658 &	56,528  \\
\#Errs	         &	1,097   &	54,173  &	19,374  &	1,093   \\
\#Backs         &	691     &	53,573  &	19,435  &	310     \\
bPrec           &	72.94\% &	95.88\% &	95.64\% &	58.06\% \\
bRec            &	45.94\% &	94.83\% &	95.94\% &	16.47\% \\
C$\rightarrow$C &	24.31\% &	03.77\% &	04.21\% &	39.35\% \\
E$\rightarrow$E &	27.06\% &	07.35\% &	05.40\% &	29.03\% \\
C$\rightarrow$E &	02.75\% &	00.35\% &	00.16\% &	02.58\% \\
E$\rightarrow$C &	45.88\% &	88.53\% &	90.23\% &	29.03\% \\ \bottomrule
\end{tabular}
\caption{Statistics for \back{} actions performed during tagging for four languages.}
\label{tab:stats_tagger_multi}
\end{table}

The statistics on the situations in which \back{} actions are performed have been displayed for four languages in Table~\ref{tab:stats_tagger_multi}. The table reveals some striking differences for two languages: German and Russian. For these languages, the ratio of \back{} actions with respect to the total number of actions predicted, is equal to $8.3\%$ for German and to $8.2\%$ for Russian, a figure that is far above what is observed for other languages. A closer look at the results shows that for these two languages, the machine learns a strategy that consists in provoking errors (it tags as punctuation linguistic tokens), in order to be able to correct them using a \back{} action. This behavior is not due to linguistic reasons but rather to the size of the training corpora. As one can see in Table~\ref{tab:corpora}, the training corpora for German and Russian are much larger than they are for other languages. Our hypothesis is that, when trained on a large training corpus, the machine has more opportunities to develop complex strategies such as provoking errors in order to correct them using a \back{} action. Indeed, this phenomenon vanishes when we reduce the size of the train set. In order to fight against this behavior, one can act on the reward function and decrease the reward of \back{} action as the size of the training corpus increases. More investigations is needed to fully understand this behavior.

The reason why the tagger chose to make regular mistakes: tagging as punctuation the words that are later corrected, is not clear. Our hypothesis is that this is an error that is easy to detect in order to predict a \back{} action.

The same phenomenon (intentional errors) has been observed, to a lesser degree, on the parser.

\section{Conclusions}
\label{sec:conclusions}

We have proposed, in this article, an extension to transition based parsing that consists in allowing the parser to undo the last actions performed in order to explore alternative hypotheses. We have shown that this kind of model can be effectively trained using deep reinforcement learning. 

This work will be extended in several ways.

The first one concerns the disappointing results of the tagparser. As already mentioned in Section~\ref{sec:training}, many studies have shown that there is usually an advantage to jointly predict several types of linguistic annotations over predicted them separately. We did not observe this phenomenon in our backtracking tagparser. The problem could come from the structure of the backtracking machines. The structures used, illustrated in Figure~\ref{fig:back-architecture} are just one possible architecture, others are possible, as for example dedicating different \back{} states for the parser and the tagger.

The second direction concerns the integration of other \nlp{} tasks in a single machine. \citet{dary-nasr-2021-reading} showed that reading machines can take as input raw text and realize word segmentation, \pos{} tagging, lemmatization, parsing and sentence segmentation. We would like to train such complex machine with reinforcement learning and backtracking to study whether the machine can backtrack across several linguistic levels.

A third direction concerns the processing of garden path sentences. Such sentences offer examples in which we expect a backtracking machine to backtrack. We have built a corpus of 54 garden path sentences in French, using four different syntactic patterns of different complexity level and organized the sentences in minimal pairs and have tested our machine on it. The results are mixed, in some cases, the machine behaves as expected, in other cases it does not. A detailed analysis showed that the machine usually backtracks on garden path sentences but has a tendency of not reanalysing the sentence as was expected. The problem seems to be of a lexical nature. Some words that should be attributed new \pos{} tags in the reanalysis phase resist this reanalysis. This is probably due to their lexical representation. More work is needed to understand this phenomenon and find ways to overcome it.

The last direction is linked to the long term project of predicting regressive saccades. The general idea is to compare back movements predicted by our model and actual eye movement data and study whether these movements are correlated.

\section*{Appendix A: Time Complexity of Backtracking}
\label{sec:complexity}

In arc-eager dependency parsing, a sentence of length $n$ is processed in $2n$ actions: an action that pushes a word on the stack (shift or right) and an action that removes it from the stack (reduce or left).
In our backtracking tagparser, we must add one tagging action and one \actionNoBack{} per word.

Therefore, without applying any \actionBack{} action, the number of actions needed to process a sentence of size $n$ is $4n$: for each word a \actionNoBack{}, a \pos{} action, a push action and a pop action.

Let $s_i$ be the length of the action sequence taking place when processing the $i^\text{th}$ word of the sentence.
The sum of these lengths is also the number of actions to process the whole sentence, therefore $\sum_{i=1}^{n}{s_i} = 4n$.

Now, in the worst case scenario where \actionBack{} is applied $k$ times per word, the total number of actions is $5nk + 4n$, which is the sum of:

\begin{flushleft}
\tabcolsep=0.1cm
\begin{tabular}{ll@{}}
~~$\bullet$ & $k$ back actions per word: $nk$.\\
~~$\bullet$ & Initial application of the $s_i$ sequences: $4n$.\\
~~$\bullet$ & The $k$ re-processing of the sequences:\\
          & $\sum_{i=1}^{n}{k\times s_i} = k\sum_{i=1}^{n}{s_i} = 4nk$.
\end{tabular}
\end{flushleft}

\section*{Appendix B: Features of the Classifiers}

The input layer of the classifiers described in Section~\ref{sec:experiments} is a vector of features extracted from the current configuration. Features are represented as randomly initialized and learnable embeddings of size $128$, with the exception of words, that are represented by fastText pretrained word embeddings of size $300$ \citep{bojanowski-etal-2017-enriching}. $4$ embedding spaces are used: words, {\sc pos} tags, letters and actions.

The features are the following:

\noindent
$\bullet$ {\sc pos} tags and form of the words in a window of size [-2,2] centered on the current word, with the addition, for the parser and tagparser, of the {\sc pos} tag of the governor, the rightmost and the leftmost dependents of the three topmost stack elements.\\
$\bullet$ The history of the $10$ last actions performed.\\
$\bullet$ Prefix and suffix of size 4 for the current word.\\
$\bullet$ For backtracking machines, a binary feature indicating whether or not a back is allowed.

When the value of a feature is not available, it is replaced by a special learnable embedding, representing the reason of unavailability. The following situation are distinguished:

\noindent
$\bullet$ Out of bounds: target word is before the first or after the last word of the sentence.\\
$\bullet$ Empty stack: target word is in the empty stack.\\
$\bullet$ No dep / gov: target word is the dependent / governor of a word without one.\\ 
$\bullet$ Not seen: target word is in right context, and has not been seen yet.\\
$\bullet$ Erased: target value has been erased after a \back{} action.

\FloatBarrier{}
\section*{Acknowledgments}

We would like to thank the action editor as well as the anonymous reviewers, for their detailed and thoughtful insights, that helped us improve on our work substantially.
\bibliography{biblio}
\bibliographystyle{acl_natbib}

\end{document}